\PassOptionsToPackage{table}{xcolor}
\documentclass[sigconf]{acmart}
\usepackage{xcolor}
\usepackage{svg}
\usepackage{bm}
\usepackage{enumitem}

\usepackage{amssymb}

\usepackage{fancyvrb}

\AtBeginDocument{%
  }

\begin{document}

\pagenumbering{gobble}
\setcounter{page}{0}

\section*{ACM Copyright Statement}

\noindent
Copyright \textcopyright 2025 held by the owner/author(s). Publication rights licensed to ACM. Permission to make digital or hard copies of all or part of this work for personal or classroom use is granted without fee provided that copies are not made or distributed for profit or commercial advantage and that copies bear this notice and the full citation on the first page. Copyrights for components of this work owned by others than the author(s) must be honored. Abstracting with credit is permitted. To copy otherwise, or republish, to post on servers or to redistribute to lists, requires prior specific permission and/or a fee. Request permissions from permissions@acm.org.

\vspace{3em}
\noindent\textbf{Published in:} Proceedings of the 19th ACM Conference on Recommender Systems, 2025 (\textit{RecSys'25, Prague, Czech Republic})

\vspace{3em}
\noindent\textbf{Cite as:}
\vspace{0.5em}

\noindent
Pedro R. Pires, Gregorio F. Azevedo, Pietro L. Campos, Rafael T. Sereicikas, and Tiago A. Almeida. 2025. Exploitation Over Exploration: Unmasking the Bias in Linear Bandit Recommender Offline Evaluation. In \textit{Proceedings of the 19th ACM Conference on Recommender Systems} (Prague, Czech Republic) \textit{(RecSys'25)}. Association for Computing Machinery, New York, NY, USA, 736-745 doi:10.1145/3705328.3748166

\vspace{3em}
\noindent\textbf{BibTeX:}
\vspace{0.5em}

\begin{Verbatim}[frame=single]
@inproceedings{10.1145/3705328.3748166,
    title       =   {Exploitation Over Exploration: Unmasking the Bias in Linear Bandit Recommender Offline
                     Evaluation},
    author      =   {Pedro R. Pires and Gregorio F. Azevedo and Pietro L. Campos and Rafael T. Sereicikas and 
                     Tiago A. Almeida},
    booktitle   =   {Proceedings of the 19th ACM Conference on Recommender Systems},
    series      =   {RecSys'25},
    location    =   {Prague, Czech Republic},
    publisher   =   {Association for Computing Machinery},
    address     =   {New York, NY, USA},
    pages       =   {736--745},
    year        =   {2025},
    doi         =   {10.1145/3705328.3748166},
    keywords    =   {Contextual Multi-Armed Bandits, Linear Bandits, Offline Evaluation, Exploration, 
                     Exploitation, Offline Recommender Systems},
}
\end{Verbatim}

\newpage
\pagenumbering{arabic}

\title{Exploitation Over Exploration: Unmasking the Bias in Linear Bandit Recommender Offline Evaluation}


\author{Pedro R. Pires}
    \orcid{0000-0001-7990-9097}
    \affiliation{%
        \institution{Federal University of São Carlos}
        \city{Sorocaba} 
        \state{São Paulo} 
        \country{Brazil}
    }
    \email{pedro.pires@dcomp.sor.ufscar.br}

\author{Gregorio F. Azevedo}
    \orcid{0000-0002-1096-7456}
    \affiliation{%
        \institution{Federal University of São Carlos}
        \city{Sorocaba} 
        \state{São Paulo} 
        \country{Brazil}
    }
    \email{gregorio.fornetti@estudante.ufscar.br}

\author{Pietro L. Campos}
    \orcid{0009-0004-3378-5921}
    \affiliation{%
        \institution{Federal University of São Carlos}
        \city{Sorocaba} 
        \state{São Paulo} 
        \country{Brazil}
    }
    \email{pietro.campos@estudante.ufscar.br}

\author{Rafael T. Sereicikas}
    \orcid{0009-0009-9198-5469}
    \affiliation{%
        \institution{Federal University of São Carlos}
        \city{Sorocaba} 
        \state{São Paulo} 
        \country{Brazil}
    }
    \email{rafaeltofoli@estudante.ufscar.br}

\author{Tiago A. Almeida}
    \orcid{0000-0001-6943-8033}
    \affiliation{%
        \institution{Federal University of São Carlos}
        \city{Sorocaba} 
        \state{São Paulo} 
        \country{Brazil}
    }
    \email{talmeida@ufscar.br}

\renewcommand{\shortauthors}{Pires et al.}

\begin{abstract}
  Multi-Armed Bandit (MAB) algorithms are widely used in recommender systems that require continuous, incremental learning. A core aspect of MABs is the exploration–exploitation trade-off: choosing between exploiting items likely to be enjoyed and exploring new ones to gather information. In contextual linear bandits, this trade-off is particularly central, as many variants share the same linear regression backbone and differ primarily in their exploration strategies. Despite its prevalent use, offline evaluation of MABs is increasingly recognized for its limitations in reliably assessing exploration behavior. This study conducts an extensive offline empirical comparison of several linear MABs. Strikingly, across over 90\% of various datasets, a greedy linear model---with no type of exploration---consistently achieves top-tier performance, often outperforming or matching its exploratory counterparts. This observation is further corroborated by hyperparameter optimization, which consistently favors configurations that minimize exploration, suggesting that pure exploitation is the dominant strategy within these evaluation settings. Our results expose significant inadequacies in offline evaluation protocols for bandits, particularly concerning their capacity to reflect true exploratory efficacy. Consequently, this research underscores the urgent necessity for developing more robust assessment methodologies, guiding future investigations into alternative evaluation frameworks for interactive learning in recommender systems. The source code for our experiments is publicly available on \url{https://github.com/UFSCar-LaSID/exploit-over-explore}.
\end{abstract}




\maketitle

\section{Introduction}

Recommender systems must adapt to evolving user preferences in dynamic environments. Contextual Multi-Armed Bandits (CMABs) with linear models have become a standard approach for sequential recommendation, leveraging user and item features to guide adaptive selection~\citep{LiContextual2010,SilvaMulti2022}. Central to CMABs is the exploration-exploitation trade-off: balancing immediate rewards from known items against exploration of uncertain options to improve long-term performance. This exploration capability is vital for fostering content discovery and maintaining catalog diversity in real-world applications~\citep{ChenValue2021}.

Despite the theoretical advantages of MABs, particularly their exploratory capabilities, evaluating bandit algorithms remains a substantial challenge. Two primary methodologies exist: \textbf{online evaluation}, involving live A/B tests with real user interactions, and \textbf{offline evaluation} (often termed replay evaluation~\citep{LiContextual2010}), which leverages historical log data. While online testing provides the most definitive measure of an algorithm's real-world performance, its associated costs, implementation complexities, and potential risks to user experience render it impractical for routine algorithm development and preliminary research. Consequently, offline evaluation remains the \textit{de facto} standard for benchmarking MAB algorithms. However, this approach is inherently passive; the logged rewards are fixed, and the algorithm's hypothetical recommendations do not influence the observed user behavior~\citep{DudikDoubly2011,LiOffline2018}. These counterfactual limitations are particularly visible when assessing exploration strategies, as their performance critically depends on observing outcomes from actions that the logging policy may not have taken.

This paper directly confronts this methodological challenge, contending that current offline evaluation practices for linear CMABs in recommender systems are not merely limited but are systematically biased against exploration. Through an extensive empirical study comparing traditional linear CMABs against their purely greedy counterpart, we reveal a counterintuitive yet persistent phenomenon: the greedy strategy consistently achieves top-tier performance. This outcome suggests that the very design of common offline protocols disincentivizes or even penalizes exploration. This bias is further evidenced during hyperparameter optimization, where configurations that minimize exploration are frequently selected. Our findings thus unmask an inherent predisposition within these evaluation settings, where the perceived benefits of sophisticated exploration strategies are largely negated.

Consequently, we argue for a critical re-examination of offline evaluation standards for MAB-based recommenders. This includes a discussion of alternative approaches, such as the creation of semi-synthetic or simulated environments that allow for a more faithful assessment of exploration, and the use of datasets specifically curated for less biased offline evaluation.

In summary, our main contributions are:
\begin{itemize}[leftmargin=*, itemsep=2pt]
    \item[\textit{(i)}] We conduct a large-scale empirical comparison of prominent linear contextual bandit algorithms under widely-used offline evaluation protocols, demonstrating the surprising efficacy of purely exploitative models;
    \item[\textit{(ii)}] We reveal and quantify a systematic bias against exploration inherent in these offline protocols, showing how exploration mechanisms are consistently devalued, leading to the artificial dominance of greedy strategies during hyperparameter tuning;
    \item[\textit{(iii)}] We advocate for a critical reassessment of offline MAB evaluation practices, discussing the potential and current limitations of alternative methodologies, including simulation-based approaches, and outlining crucial directions for future research to foster more reliable bandit algorithm assessment.
\end{itemize}

\section{Related Work}

The importance of continuously adapting to evolving user preferences has positioned Multi-Armed Bandit (MAB) algorithms as a cornerstone of modern incremental recommender systems~\citep{VinagreOverview2015,GamaSurvey2014}. Contextual MABs (CMABs) enhance these algorithms by incorporating side information (user/item features) into the decision-making process while training an underlying model. One of the most traditional approaches is the use of a linear backbone, in which rewards are modeled as a linear function of these features, with influential variants like LinGreedy~\citep{LangfordEpoch2007}, LinUCB~\citep{LiContextual2010}, and LinTS~\citep{AgrawalThompson2013} differing primarily in their strategies for navigating the fundamental exploration-exploitation trade-off. This trade-off dictates balancing the exploitation of known high-reward items against the exploration of novel or uncertain options to gather new information, a critical process for avoiding local optima and ensuring long-term performance~\citep{SlivkinsIntroduction2019}.

While much research focuses on optimizing reward prediction for exploitation, the strategic value of exploration extends beyond mere accuracy. Users often engage with recommender systems not just for precise predictions of established tastes, but to discover novel items within diverse catalogs~\citep{SilvaMulti2022,ChenValue2021,ParaparUnified2021}. Indeed, exploration mechanisms can significantly enrich user experience and have been shown to subtly influence user discovery paths and decision-making processes~\citep{LiCascading2020,WangIncentivizing2021,Caraban23Ways2019,JesseDigital2021,LiangRole2021,LiangExploring2022}.

Evaluating CMABs remains challenging. While online evaluation provides strong guarantees, its cost and risk make it impractical for most settings~\citep{GilotteOffline2018,SilvaMulti2022}. As a result, offline evaluation with logged data is widely used~\citep{AkkerPractical2023}, despite its known biases when assessing counterfactual or exploratory behavior~\citep{GuptaOptimal2024,YangUnbiased2018}.

To mitigate these biases, various off-policy evaluation (OPE) techniques have been proposed, including Inverse Propensity Scoring~\citep{WangOptimal2016}, Doubly Robust estimators~\citep{DudikDoubly2011}, and Replayers~\citep{LiUnbiased2011,ZengOnline2016,TangEnsemble2014}. These methods attempt to correct for discrepancies between the logging and target policies, yet they grapple with the bias-variance trade-off: aggressive bias reduction often inflates variance, yielding unreliable estimates~\citep{GuptaOptimal2024}. Simulation environments~\citep{RohdeRecoGym2018,IeRecSim2019,ZhaoKuaiSim2023} offer another avenue, allowing controlled experimentation with arbitrary policies and evaluation of aspects like robustness to concept drift~\citep{CaropreseModelling2025}. These simulators can avoid some counterfactual limitations of static logs~\citep{LiCascading2020,WangIncentivizing2021,ZhuDeep2023}, but building high-fidelity simulators that accurately reflect complex real-world user behavior remains a very challenging task.

Despite the availability of advanced OPE methods and simulators, naive offline replay remains the dominant practice for evaluating CMABs, even for exploratory algorithms~\citep{SilvaMulti2022}. This reliance raises concerns about the validity of such evaluations, especially for \emph{linear contextual bandits}, whose effectiveness depends on interaction-driven exploration. Offline protocols often favor exploitative behavior, obscuring the long-term value of exploration.

To investigate this, we conduct a large-scale offline comparison between standard exploratory linear bandits and a purely greedy linear model---the common backbone for many CMABs. In 89\% of evaluated datasets, the greedy model matched or outperformed its exploratory counterparts. This counterintuitive finding highlights how standard offline evaluation can systematically penalize exploration, misleading conclusions about algorithmic effectiveness. To our knowledge, this is the first study to quantify this bias at scale for linear bandits; while~\citet{RaoContextual2020} presented related observations, his study was conducted on a significantly smaller scale and with a different analytical focus.

\section{Linear Contextual Bandit Algorithms}
\label{sec:linear_bandits}

Multi-Armed Bandit (MAB) algorithms traditionally observe rewards only for selected arms. Contextual MABs (CMABs) extend this by incorporating context—typically feature vectors—allowing agents to adapt decisions to each situation. This brings CMABs closer to supervised learning while preserving their interactive, sequential nature. A prominent subclass is linear contextual bandits, which use linear regression models for reward prediction.

This section details the linear CMAB algorithms evaluated in our study, emphasizing their shared linear regression foundation and their distinct approaches to exploration.

\subsection{Problem Formulation}
\label{subsec:problem}

In the CMAB setting, at each round $t=1, 2, \dots, T$, the agent observes a $d$-dimensional context vector $\{\mathbf{x}_{t,a} \in \mathbb{R}^d\}_{a\in\mathcal{A}}$ for each arm $a$ in a finite set of arms $\mathcal{A}$. After selecting an arm $a_t \in \mathcal{A}$, the learner observes a reward $r_{t,a_t} \in \mathbb{R}$. The objective is to devise a policy $\pi$ that selects arms to maximize the cumulative reward $\sum_{t=1}^T r_{t,a_t}$ (or, equivalently, minimize cumulative regret). The policy can leverage observed contexts, enabling personalized or item-aware decisions crucial in applications like recommender systems.

Linear contextual bandits posit that the expected reward for arm $a$ given its context $\mathbf{x}_{t,a}$ is a linear function:
\begin{equation}
    \mathbb{E}[r_{t,a} \mid \mathbf{x}_{t,a}] = \mathbf{x}_{t,a}^\top \bm{\theta}_a,
    \label{eq:linear_model}
\end{equation}
where the unknown parameter vector $\bm{\theta}_a \in \mathbb{R}^d$ is specific to each arm $a$. This arm-specific parameterization decouples estimation across arms, preventing interference and allowing for heterogeneous reward structures.

\subsection{Incremental Per-Arm Linear Models}
\label{subsec:updates}

The parameters $\bm{\theta}_a$ are typically estimated using regularized least-squares (ridge regression) independently for each arm. For each arm $a \in \mathcal{A}$, we maintain a $d \times d$ matrix $\mathbf{A}_a$ and a $d$-dimensional vector $\mathbf{b}_a$. Initially, $\mathbf{A}_a = \lambda\mathbf{I}_d$ and $\mathbf{b}_a = \mathbf{0}_d$, where $\lambda > 0$ is the regularization parameter and $\mathbf{I}_d$ is the $d \times d$ identity matrix.

When arm $a_t$ is selected at round $t$ with context $\mathbf{x}_{t,a_t}$ and observed reward $r_{t,a_t}$, the statistics for arm $a_t$ are updated as follows:
\begin{align}
    \mathbf{A}_{a_t} &\leftarrow \mathbf{A}_{a_t} + \mathbf{x}_{t,a_t}\mathbf{x}_{t,a_t}^\top, \\
    \mathbf{b}_{a_t} &\leftarrow \mathbf{b}_{a_t} + r_{t,a_t}\mathbf{x}_{t,a_t}.
\end{align}

The estimate of the parameter vector for arm $a$ is then given by:

\begin{equation}
    \hat{\bm{\theta}}_a = \mathbf{A}_a^{-1}\mathbf{b}_a.
    \label{eq:theta_hat}
\end{equation}

These estimates $\hat{\bm{\theta}}_a$ can be updated efficiently using the Sherman-Morrison formula for rank-one updates to $\mathbf{A}_a^{-1}$, facilitating real-time learning.

\subsection{Action-Selection Strategies}
\label{subsec:strategies}

The core linear bandits differ in how they use the estimates $\hat{\bm{\theta}}_a$ and the uncertainty information encapsulated in $\mathbf{A}_a$ to select an arm.

\textbf{Notation Convention.} To ensure clarity, we adopt the following nomenclature:
\begin{itemize}[leftmargin=*]
    \item \textbf{Lin}: The purely greedy policy, equivalent to $\varepsilon$-greedy with $\varepsilon=0$. It always exploits the current best estimate.
    \item \textbf{LinGreedy}: The $\varepsilon$-greedy policy. When $\varepsilon > 0$, it introduces random exploration. When $\varepsilon=0$, it is identical to Lin.
\end{itemize}

\paragraph{\textbf{Lin} (Purely Greedy).}
This strategy selects the arm with the highest predicted reward based on the current estimates:
\begin{equation}
    a_t = \operatorname*{arg\,max}_{a\in\mathcal{A}}\; \mathbf{x}_{t,a}^\top\hat{\bm{\theta}}_a.
    \label{eq:lin}
\end{equation}
No explicit exploration mechanism is employed; any exploration is incidental, primarily arising from initial model uncertainty before the estimates $\hat{\bm{\theta}}_a$ stabilize.

\paragraph{\textbf{LinGreedy} ($\varepsilon$-Greedy)~\citep{LangfordEpoch2007}.}
This policy explores by selecting an arm uniformly at random from $\mathcal{A}$ with probability $\varepsilon \in [0,1]$ and exploits by choosing the greedy arm with probability $1-\varepsilon$:
\begin{equation}
    a_t =
    \begin{cases}
      \displaystyle\operatorname*{arg\,max}_{a\in\mathcal{A}}\; 
      \mathbf{x}_{t,a}^\top\hat{\bm{\theta}}_a, & \text{with probability } 1-\varepsilon,\\[6pt]
      \text{select uniformly from }\mathcal{A}, & \text{with probability } \varepsilon.
    \end{cases}
    \label{eq:epsilon_lin}
\end{equation}

\paragraph{\textbf{LinUCB}~\citep{LiContextual2010}.}
The Linear Upper Confidence Bound algorithm incorporates an exploration bonus that encourages selecting arms with high uncertainty. It selects $a_t = \operatorname*{arg\,max}_{a\in\mathcal{A}} \; p_{t,a}$, where the score $p_{t,a}$ is calculated as:
\begin{equation}
    p_{t,a} = \mathbf{x}_{t,a}^\top\hat{\bm{\theta}}_a
    + \alpha\,\sqrt{\mathbf{x}_{t,a}^\top\mathbf{A}_a^{-1}\mathbf{x}_{t,a}}.
    \label{eq:linucb}
\end{equation}
Here, $\alpha \ge 0$ is a hyperparameter controlling the degree of exploration, and the second term serves as an upper confidence bound on the reward estimate, promoting exploration of arms whose context vectors align with directions of higher parameter uncertainty.

\paragraph{\textbf{LinTS}~\citep{AgrawalThompson2013}.}
Linear Thompson Sampling applies Bayesian principles to exploration. At each round $t$, for each arm $a$, it samples a parameter vector $\tilde{\bm{\theta}}_a$ from its current posterior distribution over $\bm{\theta}_a$. Assuming a Gaussian likelihood for rewards, the posterior (or an approximation under ridge regression) is also Gaussian:
\begin{equation}
    \tilde{\bm{\theta}}_a \sim 
        \mathcal{N}\!\bigl(\hat{\bm{\theta}}_a, \nu^{2}\mathbf{A}_a^{-1}\bigr).
    \label{eq:lints_sample}
\end{equation}
The arm $a_t$ is then chosen greedily with respect to these sampled parameters:
\begin{equation}
    a_t=\operatorname*{arg\,max}_{a\in\mathcal{A}}\; \mathbf{x}_{t,a}^\top\tilde{\bm{\theta}}_a.
    \label{eq:lints_select}
\end{equation}

The hyperparameter $\nu^2 > 0$ (often related to an assumed noise variance or a user-defined scaling factor) controls the amount of exploration. This probability-matching approach naturally balances exploration-exploitation based on current parameter uncertainty.

All four algorithms leverage the same incremental per-arm linear model structure (Equations~\eqref{eq:linear_model}--\eqref{eq:theta_hat}). Their differentiation lies entirely in the strategy used to convert these learned estimates and associated uncertainties into arm selections.

\section{Datasets for Offline Evaluation}

Offline evaluation using historical interaction logs remains the standard approach for assessing MAB algorithms in recommender systems~\citep{SilvaMulti2022}. Typically, existing recommendation datasets are adapted to the bandit setting by simulating incremental learning over time-ordered user interactions. However, the lack of consensus on dataset selection and preprocessing protocols poses challenges for reproducibility, complicating fair comparisons and meta-analyses.

Table~\ref{tab:offline_datasets} presents a compilation of datasets frequently employed in the MAB literature for offline evaluation, alongside representative studies that utilize them. This highlights the diversity in data sources and domains, but also underscores the varied landscape within which bandit algorithms are benchmarked.

\begin{table}[htbp]
    \centering
    \caption{List of datasets commonly used for offline multi-armed bandits evaluation.}\label{tab:offline_datasets}
    \begin{tabular}{lccl}
        \textbf{Dataset} & \textbf{Public} & \textbf{Domain} & \textbf{Papers}  \\ \toprule
        \textbf{Amazon Review} & \checkmark & Retail & \citep{ChenKnowledge2020,GhoorchianNonStationary2024} \\
        \textbf{Book-Crossing} & \checkmark & Books & \citep{ZhouInteractive2020} \\
        \textbf{Cheetah Mobile} & & Articles & \citep{LiuTransferable2018} \\
        \textbf{Delicious} & \checkmark & Bookmarks & \citep{BianchiGang2013,LiuTransferable2018,JagermanWhen2019,NguyenDynamic2014,WangLearning2016} \\
        \textbf{GoodReads} & \checkmark & Books & \cite{ChenKnowledge2020} \\
        \textbf{Jester} & \checkmark & Jokes & \citep{GhoorchianNonStationary2024} \\
        \textbf{Last.FM} & \checkmark & Music & \citep{BianchiGang2013,CaronMixing2013,JagermanWhen2019,NguyenDynamic2014,WangLearning2016,WangFactorization2017,XuContextual2020} \\
        \textbf{Million Songs} & \checkmark & Music & \citep{TakemoriSubmodular2020} \\
        \textbf{MovieLens} & \checkmark & Movies & \citep{CelisControlling2019,ChenKnowledge2020,GhoorchianNonStationary2024,RaoContextual2020,TakemoriSubmodular2020,ZhouInteractive2020} \\
        \textbf{NetflixPrize} & \checkmark & Movies & \citep{ChenKnowledge2020} \\
        \textbf{PoliticalNews} & & News & \citep{CelisControlling2019} \\
        \textbf{Spotify} & & Music & \citep{McInerneyExplore2018} \\
        \textbf{Toutiao} & \checkmark & News & \citep{ZhangConversational2020} \\
        \textbf{Xiami Music} & & Music & \citep{ZhouConversational2020} \\
        \textbf{Yahoo! Music} & \checkmark & Music & \citep{HaririAdapting2015} \\
        \textbf{Yahoo! News} & \checkmark & News & \citep{ChapelleEmpirical2011,SongInteractions2019,TracaReducing2019,WangFactorization2017,WuReturning2017,XuContextual2020} \\
        \textbf{Yelp} & \checkmark & Restaurants & \citep{ChenContextual2018,ZhangConversational2020} \\
        \textbf{YOW} & \checkmark & News & \citep{CelisControlling2019} \\ 
        \bottomrule
    \end{tabular}
\end{table}

When timestamped interactions are available, a common approach involves sorting data chronologically to simulate a sequential process, often framed as a next-item prediction task~\citep{HaririAdapting2015,TracaReducing2019,ZhouConversational2020}. To manage computational complexity, or adapt to specific goals, researchers often apply preprocessing steps, e.g., subsampling popular items or active users~\citep{CaronMixing2013,TakemoriSubmodular2020,WuReturning2017,ZhouInteractive2020}; restricting the candidate available arm set~\citep{BianchiGang2013,JagermanWhen2019,LiuTransferable2018,NguyenDynamic2014}; or altering recommendation granularity through user or item clustering~\citep{CelisControlling2019,WangFactorization2017}.

While such preprocessing decisions are often practical necessities, their considerable variability across studies, coupled with the absence of a broadly adopted evaluation protocol, exacerbates the difficulty of rigorously assessing MAB algorithms~\cite{SilvaMulti2022}. This lack of methodological consistency is particularly detrimental for evaluating exploration strategies, as many \textit{ad-hoc} choices can inadvertently influence outcomes and obscure the benefits of nuanced exploration. To address this, we apply a unified and reproducible offline evaluation protocol to a suite of contextual linear bandits, as explained in the following sections.

\section{Experimental Setup}
In this section, we delineate the employed experimental protocol. We first describe the datasets used, followed by a detailed account of the benchmarking procedure.

\subsection{Datasets}
Our experiments were conducted on a diverse collection of publicly available datasets, commonly employed in recommender systems and bandit algorithm research. These are summarized in Table~\ref{tab:datasets}. During preprocessing, we removed duplicate interactions and discarded inconsistent entries, i.e., same user–item pair with conflicting ratings. Finally, for dataset \textit{Delicious}, we generated an additional dataset, \textit{Delicious-PU}, containing only the principal URL.

\begin{table}[htpb]
\centering
\caption{Datasets used in the experiments.}
\label{tab:datasets}
\begin{tabular}{lcccc}
\toprule
\textbf{Dataset} & \textit{\#users} & \textit{\#items} & \textit{\#interactions}\\
\midrule
\textbf{Amazon Beauty}\footnotemark[1]          &    631,986 &  112,565 &  701,528\\
\textbf{Amazon Books}\footnotemark[1]        &  1,008,954 &  206,710 &  2,437,999\\
\textbf{Amazon Games}\footnotemark[1]        &  2,766,656 &  137,249 &  4,624,615\\
\textbf{BestBuy}\footnotemark[2]        &  1,268,702 &  69,858 &  1,865,269\\
\textbf{Delicious}\footnotemark[3]        &  1,867 &  69,198 &  437,593\\
\textbf{MovieLens-100K}\footnotemark[4]        &    943 &  1,682 &    100,000\\
\textbf{MovieLens-25M}\footnotemark[4]      &     162,541 &  59,047 &   25,000,095\\
\textbf{RetailRocket}\footnotemark[5]      &     1,407,580 &   235,061 &    2,755,641\\
\bottomrule
\end{tabular}
\end{table}

\footnotetext[1]{Amazon Reviews'23. Available at: \url{https://amazon-reviews-2023.github.io} (visited on \today)}
\footnotetext[2]{Data Mining Hackathon on BIG DATA (7GB). Available at: \url{www.kaggle.com/c/acm-sf-chapter-hackathon-big} (visited on \today)}
\footnotetext[3]{HetRec2011 | GroupLens. Available at: \url{https://grouplens.org/datasets/hetrec-2011/} (visited on \today)}
\footnotetext[4]{MovieLens | GroupLens. Available at: \url{https://grouplens.org/datasets/movielens/} (visited on \today)}
\footnotetext[5]{Retailrocket recommender system dataset. Available at \url{https://www.kaggle.com/datasets/retailrocket/ecommerce-dataset} (visited on \today)}

\subsection{Experimental Protocol}

\begin{figure*}
    \centering
    \includegraphics[width=0.65\linewidth]{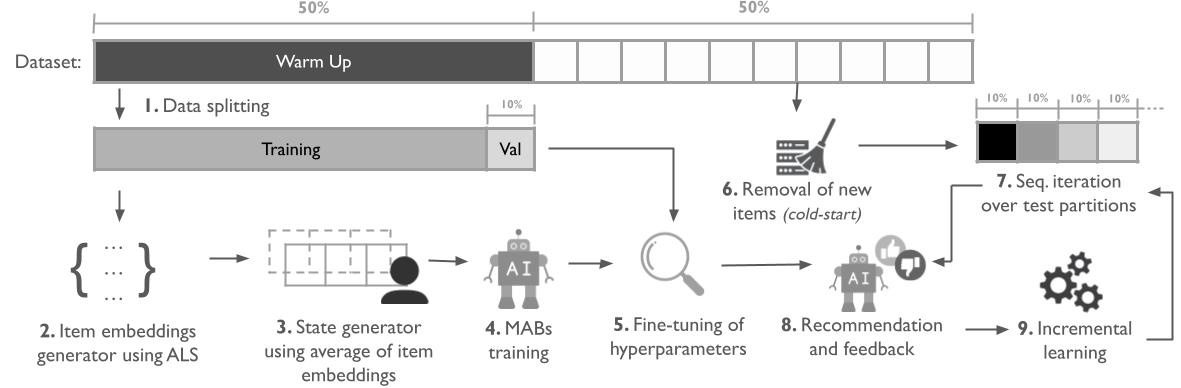}
    \caption{Experimental protocol used to benchmark the models.}
    \label{fig:experimental-protocol}
    \Description[Diagram of experimental protocol]{Black and white sequential diagram showing the adopted experimental protocol. The pipeline is divided into 9 steps, each with a corresponding icon, illustrating what was described in the text.}
\end{figure*}

To evaluate the models, we followed the pipeline illustrated in Figure~\ref{fig:experimental-protocol}. Each dataset was chronologically sorted to simulate a continuous environment. The first 50\% of interactions were used for training (\textit{warm-up}), with 10\% of this portion held out for validation. The remaining 50\% was used for testing.

Item embeddings were trained on the warm-up data using ALS~\citep{HuCollaborative2008}, and user states were computed as the average of previously consumed item embeddings. These vectors served as context for training the linear CMABs. Hyperparameters for ALS---latent dimensions (${32, 64, 128}$), regularization (${32, 64, 128}$), and iteration count (${5, 15, 30}$)---were selected based on the NDCG in a top-20 recommendation setting. Since embeddings were pre-trained non-incrementally, test interactions involving unseen items were excluded.

The test partition was then divided into 10 sequential batches, each with 10\% of the data. At each batch, agents generated recommendations, received user feedback, and updated their linear models incrementally.

The entire pipeline was implemented in \texttt{Python3}, using the \texttt{implicit} library for the ALS embeddings\footnotemark[6] and \texttt{Mab2Rec}\cite{KadiogluMab2Rec2024} for the bandit algorithms.

\footnotetext[6]{Implicit -- Fast Python Collaborative Filtering for Implicit Datasets. Available at: \url{https://github.com/benfred/implicit} (visited on \today)}

\section{Experiments and Discussion}

We conducted a comprehensive set of experiments designed to address the following central research questions (RQs) regarding the offline evaluation of linear contextual bandits:

\begin{itemize}[wide, labelwidth=!,itemindent=!,labelindent=0pt, leftmargin=0em, itemsep=3pt]
    \item \textbf{RQ1:} How reliably do current offline evaluation protocols assess the efficacy of different exploration strategies in linear CMABs?
    \item \textbf{RQ2:} To what extent do purely exploitative (greedy) linear models outperform or match their exploratory counterparts under these standard offline evaluation settings?
    \item \textbf{RQ3:} What influence do hyperparameter optimization choices exert on the apparent preference for exploitation-focused strategies within offline evaluation paradigms?
\end{itemize}

We then divided the analysis into four parts: \textit{(i)} accuracy; \textit{(ii)} diversity; \textit{(iii)} hyperparameter selection; \textit{(iv)} off-policy evaluation. For transparency and reproducibility, the complete source code for our experiments is available at: \url{https://github.com/UFSCar-LaSID/exploit-over-explore}.

\subsection{Accuracy Analysis}

\begin{figure*}
    \centering
    \includegraphics[width=0.71\linewidth]{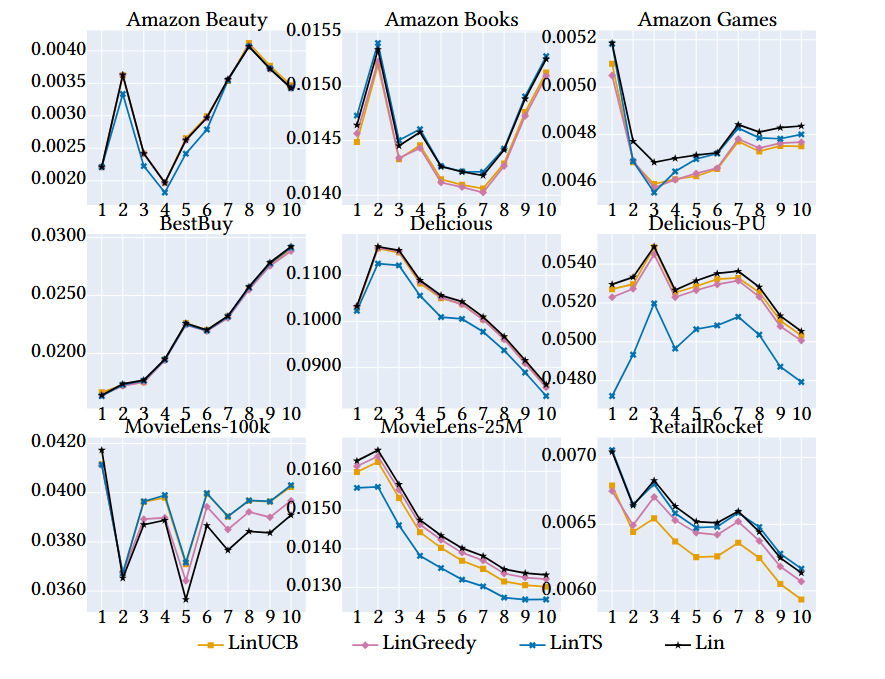}
    \caption{Cumulative NDCG@20 for every partition on the test set.}
    \label{fig:ndcg_results}
    \Description[Grid of 9 line plots for NDCG@20]{Grid with 9 subplots consisting of line plots. The x axis represents the corresponding text window, the y axis contains the NDCG for a top-20 recommendation, and the four lines show the results obtained for each algorithm. Results vary depending on the dataset, but always follow the same pattern, with the Lin results normally slightly on top.}
\end{figure*}

To evaluate the effectiveness of each model in predicting user preferences within their evolving contexts, we computed the NDCG@20 for each incremental batch in the test set. Figure~\ref{fig:ndcg_results} depicts the cumulative NDCG@20 performance as models process these sequential batches, illustrating how model accuracy evolves, and Table~\ref{tab:ndcg_grayscale} summarizes the final aggregated NDCG@20 over the entire test period. Darker cell shading indicates superior performance, and the top-ranked result for each dataset is highlighted in \textbf{bold}\footnotemark[7].

\footnotetext[7]{Additional metrics such as \textbf{Hit Rate} and \textbf{F1-Score}---whose trends are consistent with the NDCG findings---, as well as the non-aggregated results for the NDCG and different values of $K$, were also computed and are available in the project's documentation on the GitHub repository.}

The performance trajectories illustrated in Figure~\ref{fig:ndcg_results} underscore the pervasive influence of the shared linear regression backbone common to all evaluated algorithms. Across most datasets, the exploratory MAB variants exhibit curves closely tracking, and often failing to meaningfully distinguish themselves from, the purely greedy Lin model. This persistent parallelism directly addresses \textbf{RQ1} by highlighting a critical limitation of this offline evaluation paradigm. Specifically, if exploration strategies were conferring distinct advantages, we would anticipate more significant divergences in performance over time, with effective explorers potentially surpassing the greedy baseline as they accumulate more diverse information. Instead, explicit exploration often slightly degrades NDCG@20, indicating that offline replay inherently tends to reward immediate and historically validated (greedy) choices over potentially beneficial but unobserved exploratory actions.

\begin{table}[htb]
\centering
\caption{NDCG in a top-20 recommendation task (NDCG@20).}\label{tab:ndcg_grayscale}
\resizebox{0.5\textwidth} {!}{ \begin{tabular}{lcccc}
\toprule
\textbf{Dataset} & \textbf{Lin} & \textbf{LinUCB} & \textbf{LinGreedy} & \textbf{LinTS} \\
\midrule
Amazon Beauty & \cellcolor[gray]{0.8914346466955507}0.00342 & \cellcolor[gray]{0.5}\textbf{0.00346} & \cellcolor[gray]{0.8914346466955507}0.00342 & \cellcolor[gray]{0.9499999999999993}0.00341 \\
Amazon Books & \cellcolor[gray]{0.5631541760206247}0.01525 & \cellcolor[gray]{0.8678854082756757}0.01512 & \cellcolor[gray]{0.9500000000000028}0.01509 & \cellcolor[gray]{0.5}\textbf{0.01527} \\
Amazon Games & \cellcolor[gray]{0.5}\textbf{0.00483} & \cellcolor[gray]{0.9499999999999993}0.00475 & \cellcolor[gray]{0.8598568413415144}0.00477 & \cellcolor[gray]{0.6826898616392576}0.00480 \\
BestBuy & \cellcolor[gray]{0.5}\textbf{0.02923} & \cellcolor[gray]{0.7492989927221032}0.02902 & \cellcolor[gray]{0.9500000000000028}0.02885 & \cellcolor[gray]{0.5928054826225022}0.02915 \\
Delicious & \cellcolor[gray]{0.5}\textbf{0.08634} & \cellcolor[gray]{0.5685167034187852}0.08596 & \cellcolor[gray]{0.6106226319702479}0.08573 & \cellcolor[gray]{0.9500000000000011}0.08385 \\
Delicious-PU & \cellcolor[gray]{0.5}\textbf{0.05055} & \cellcolor[gray]{0.5375913072011951}0.05033 & \cellcolor[gray]{0.583339631492727}0.05006 & \cellcolor[gray]{0.9500000000000011}0.04794 \\
MovieLens-100k & \cellcolor[gray]{0.9499999999999993}0.03900 & \cellcolor[gray]{0.528632769848155}0.04013 & \cellcolor[gray]{0.7337430394257876}0.03958 & \cellcolor[gray]{0.5}\textbf{0.04021} \\
MovieLens-25M & \cellcolor[gray]{0.5}\textbf{0.01332} & \cellcolor[gray]{0.7173726287161877}0.01302 & \cellcolor[gray]{0.5779718131321303}0.01321 & \cellcolor[gray]{0.9499999999999993}0.01269 \\
RetailRocket & \cellcolor[gray]{0.5611130382801939}0.00614 & \cellcolor[gray]{0.9499999999999993}0.00594 & \cellcolor[gray]{0.6879967584292572}0.00607 & \cellcolor[gray]{0.5}\textbf{0.00617} \\
\bottomrule
\end{tabular} }
\end{table}

The aggregated NDCG@20 scores in Table~\ref{tab:ndcg_grayscale} provide compelling evidence to address \textbf{RQ2}. The purely exploitative Lin baseline achieved the highest NDCG@20 in 56\% of the datasets and secured either the top or second-best rank in 89\% of all cases, being proven statistically superior to LinGreedy through a two-tailed Bonferroni-Dunn test, with 90\% confidence ($CD = 1.29$), and ranking higher than the other CMABs on average. Exploratory algorithms surpassed Lin only marginally and infrequently: LinUCB on the Amazon Beauty dataset, and LinTS on the Amazon Books and RetailRocket datasets. Even in these exceptional instances, the performance gain was minimal. More commonly, the incorporation of explicit exploration resulted in a decrease in NDCG@20 compared to the greedy baseline under these offline conditions, reinforcing the notion that such protocols may penalize exploration.

A notable exception is MovieLens-100k, where all exploratory models outperformed the Lin baseline. This likely stems from its characteristics: a minimum user activity threshold results in denser interaction histories, and its relatively small item catalog reduces the exploration ``cost''. Together, these factors make it easier for exploratory algorithms to find relevant items within offline replay, boosting their performance.

\subsection{Diversity Analysis}

\begin{figure*}
    \centering
    \includegraphics[width=0.71\linewidth]{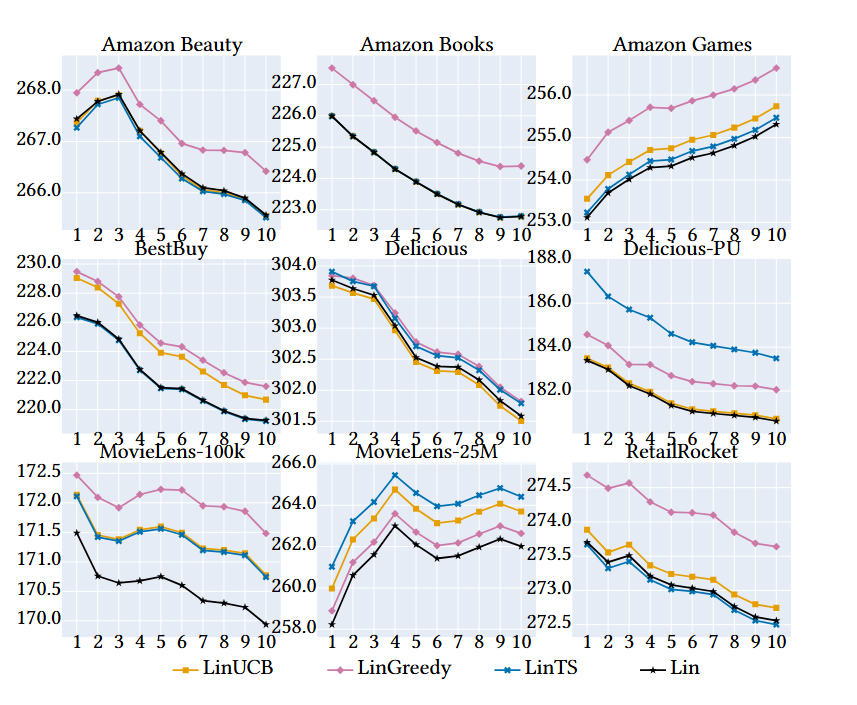}
    \caption{Cumulative novelty@20 for every partition on the test set. Higher values mean recommendations of less pop. items.}
    \label{fig:novelty_results}
    \Description[Grid of 9 line plots for novelty@20]{Grid with 9 subplots consisting of line plots. The x axis represents the corresponding text window, the y axis contains the novelty for a top-20 recommendation, and the four lines show the results obtained for each algorithm. Results vary depending on the dataset, but always follow the same pattern, with LinGreedy results normally slightly on top.}
\end{figure*}

The capacity of recommenders to generate diverse suggestions is a critical evaluation dimension. While diversity can be quantified through metrics such as coverage, serendipity, and intra-list similarity, our analysis focuses on \textbf{novelty}. Novelty quantifies the degree to which recommended items are uncommon, reflecting their divergence from typically consumed items~\citep{ShaniEvaluating2011}. We compute novelty using Equation~\ref{eq:novelty}, where $I$ represents the set of recommended items and $|R|$ denotes the number of interactions. The term $pop(i)$ is the popularity of item $i$, defined as its relative frequency of appearance within the dataset. Figure~\ref{fig:novelty_results} depicts the cumulative novelty scores, where higher values indicate more novel recommendations\footnotemark[8].

\footnotetext[8]{Additional metrics such as \textbf{Coverage}, as well as the non-aggregated results for the Novelty and different values of $K$, were also computed and are available in the project's documentation on the GitHub repository.}

\begin{equation}
    \label{eq:novelty}
    \mathrm{Novelty} = \frac{\sum_{i \in I} \log_2 pop(i)}{|R|}
\end{equation}

As expected, exploratory bandits generally yield higher novelty scores, reflecting their propensity to recommend less popular items. Nevertheless, elevated novelty alone does not guarantee superior recommendation quality; a random model, for instance, might achieve high novelty while offering largely irrelevant suggestions. This underscores the necessity of evaluating novelty in conjunction with accuracy-focused metrics. Effective recommender systems must therefore balance exploration (diversity) and exploitation (accuracy), rather than optimizing either in isolation.

A comparison of novelty and NDCG reveals an inverse relationship: models producing more diverse recommendations often show notably lower NDCG. This occurs, for example, with LinGreedy on Amazon Books and Amazon Games, and LinTS on Delicious-PU and MovieLens-25M. In these cases, increased exploration (and novelty) corresponds to reduced accuracy. These findings suggest that even with diversity metrics, standard offline evaluation protocols fail to fully capture the value of exploratory strategies, reinforcing our \textbf{RQ1} conclusions.

\subsection{Hyperparameter Selection}

To address \textbf{RQ3}, we investigated the hyperparameter configurations that maximized NDCG@20 on the validation set. For LinUCB and LinTS, exploration-controlling hyperparameters $\alpha$ and $\nu^2$ were varied across the set $\{0.10, 0.50, 1.00, 1.50, 2.00\}$. For LinGreedy, $\varepsilon$ was selected from $\{0.01, 0.05, 0.10, 0.25, 0.50\}$. In all instances, lower parameter values denote diminished exploration. The optimally selected hyperparameters are detailed in Table~\ref{tab:hyperparameters}, where darker cells indicate choices favoring exploitation.

Remarkably, across \textbf{all} evaluated scenarios, both LinUCB and LinGreedy consistently converged on hyperparameter values that minimized exploration. LinTS presented a more nuanced picture: it favored increased exploration parameters on datasets such as Amazon Books, Amazon Games, and Delicious (moderately), and more substantially on Delicious-PU and MovieLens-25M. However, when LinTS selected higher exploration settings, its NDCG@20 performance was typically poor, often ranking lowest among all models. A singular exception arose on the Amazon Books dataset, where a moderate exploration level (e.g., $\alpha=0.5$) yielded the top performance, hinting that exploration can be beneficial under specific, albeit rare, offline conditions. Nevertheless, this instance is an outlier against the prevailing trend where increased exploration degrades performance within these offline evaluation frameworks.

\begin{table}[h]
    \centering
    \caption{Selected hyperparameters during fine-tuning.}
    \label{tab:hyperparameters}
    \begin{tabular}{lrrr}
        \toprule
        \textbf{Dataset} & \textbf{LinUCB ($\alpha$)} & \textbf{LinGreedy ($\varepsilon$)} & \textbf{LinTS ($\nu^2$)} \\
        \midrule
        Amazon Beauty     & \cellcolor[gray]{0.5}0.10  & \cellcolor[gray]{0.5}0.01  & \cellcolor[gray]{0.5}0.10 \\
        Amazon Books      & \cellcolor[gray]{0.5}0.10  & \cellcolor[gray]{0.5}0.01  & \cellcolor[gray]{0.6125}0.50 \\
        Amazon Games      & \cellcolor[gray]{0.5}0.10  & \cellcolor[gray]{0.5}0.01  & \cellcolor[gray]{0.6125}0.50 \\
        BestBuy           & \cellcolor[gray]{0.5}0.10  & \cellcolor[gray]{0.5}0.01  & \cellcolor[gray]{0.5}0.10 \\
        Delicious         & \cellcolor[gray]{0.5}0.10  & \cellcolor[gray]{0.5}0.01  & \cellcolor[gray]{0.6125}0.50 \\
        Delicious-PU      & \cellcolor[gray]{0.5}0.10  & \cellcolor[gray]{0.5}0.01  & \cellcolor[gray]{0.725}1.00 \\
        MovieLens-100k    & \cellcolor[gray]{0.5}0.10  & \cellcolor[gray]{0.5}0.01  & \cellcolor[gray]{0.5}0.10 \\
        MovieLens-25M     & \cellcolor[gray]{0.5}0.10  & \cellcolor[gray]{0.5}0.01  & \cellcolor[gray]{0.725}1.00 \\
        RetailRocket      & \cellcolor[gray]{0.5}0.10  & \cellcolor[gray]{0.5}0.01  & \cellcolor[gray]{0.5}0.10 \\
        \bottomrule
    \end{tabular}
\end{table}

These hyperparameter selection outcomes reinforce the conclusion from \textbf{RQ1}: standard offline evaluation is inherently constrained in its capacity to meaningfully assess the efficacy of exploration in bandit algorithms. One alternative, still within an offline framework, is off-policy evaluation, which we explore in the following section.


\subsection{Off-Policy Evaluation}

One way to mitigate the non-interventionist limitation of full offline evaluation is through off-policy evaluation (OPE)~\cite{WangOptimal2016}. By explicitly modeling the behavior policy used during data logging, OPE enables counterfactual reasoning and can reduce the bias that typically penalizes exploratory algorithms in standard offline settings.

We evaluated the linear multi-armed bandits using the Open Bandit Pipeline (OBP)~\cite{SaitoOpenBandit2021}. The agents were benchmarked with three widely adopted OPE estimators: Inverse Propensity Weighting (IPW)~\cite{WangOptimal2016}, Direct Method (DM)~\cite{BeygelzimerOffset2016}, and Doubly Robust (DR)~\cite{DudikDoubly2011}. Results are presented in Table~\ref{tab:ope_obp}.

Surprisingly, even under conditions where the bias toward exploitation is mitigated, the purely greedy model consistently outperformed the exploratory variants. This effect is particularly evident in the IPW and DR estimators, which are designed to highlight the value of exploration. Full results and comparison against the logging policy are available in our repository.

\begin{table}[!htb]
\centering
\caption{Off-policy evaluation using the Open Bandit Pipeline.}
\begin{tabular}{lcccc}
\toprule
Estimator & Lin & LinUCB & LinGreedy & LinTS \\
\midrule
IPW & \cellcolor[gray]{0.5}\textbf{0.01637} & \cellcolor[gray]{0.8582429501084599}0.0127 & \cellcolor[gray]{0.624945770065076}0.01509 & \cellcolor[gray]{0.95}0.01176 \\
DM & \cellcolor[gray]{0.5}\textbf{0.00403} & \cellcolor[gray]{0.540909090909091}0.00401 & \cellcolor[gray]{0.9090909090909092}0.00383 & \cellcolor[gray]{0.9500000000000002}0.00381 \\
DR & \cellcolor[gray]{0.5}\textbf{0.01623} & \cellcolor[gray]{0.6265306122448981}0.01499 & \cellcolor[gray]{0.6683673469387756}0.01458 & \cellcolor[gray]{0.95}0.01182 \\
\bottomrule
\end{tabular}
\label{tab:ope_obp}
\end{table}

These findings reinforce our core claim: current offline evaluation protocols, including robust OPE estimators, remain limited in their ability to capture the true benefits of exploration. This exposes a critical methodological gap in recommender systems research: current offline evaluation protocols, including robust OPE estimators, still struggle to reflect the potential advantages of exploration. While simulators have emerged as a promising direction, they face key limitations in scope and realism. The subsequent section will delve into existing simulator-based evaluation approaches, analyzing their advantages and limitations.


\section{Simulation-based Evaluation of Bandits}

Simulators offer interactive environments where bandit algorithms can be evaluated under dynamic conditions---potentially overcoming the structural bias toward exploitation seen in offline replay. Although OPE methods can partially mitigate this bias using logged data from known policies, they cannot capture the effects of novel actions or evolving user behavior~\citep{SaitoOpenBandit2021}. Simulators, by contrast, can reward exploration in ways that static logs cannot, making them a compelling, if still maturing, alternative.

A recommender system simulator models: (i) \emph{users}, with latent or observable preferences; (ii) \emph{items}, described by features; and (iii) \emph{environment dynamics} linking recommendations to stochastic user responses. Key components include user choice functions, state-transition models, and reward generation, with some simulators also modeling item generation or ecosystem agents~\citep{IeRecSim2019}. Unlike static logs, these elements interact to produce \textit{interactive trajectories}, allowing analysis of how exploration impacts future contexts and rewards.

\begin{itemize}
  \item \textbf{Virtual-Taobao}~\citep{ShiVirtualTaobao2018}: a GAN\,+\,multi-agent imitation platform virtualizing Taobao’s search engine;
  \item \textbf{RecSim}~\citep{IeRecSim2019}: a configurable, slate-based simulation environment with modular user, item, choice, and transition models;
  \item \textbf{RecSim NG}~\citep{MladenovRecSimNG2021}: its probabilistic-programming successor, supporting gradient-based policy learning and complex multi-agent ecosystem studies;
  \item \textbf{RecoGym}~\citep{RohdeRecoGym2018}: an OpenAI Gym environment that couples organic browsing with bandit-based ad impressions;
  \item \textbf{KuaiSim}~\citep{ZhaoKuaiSim2023}: a data-driven simulator for list-wise, session-level, and cross-session recommendation tasks, featuring return-time sampling;
  \item \textbf{Concept-Drift Stream Generator}~\citep{CaropreseModelling2025}: a stream-based data generator designed to inject abrupt or progressive concept drift into synthetic interaction logs.
\end{itemize}

A common strength of these tools is their ability to \textit{execute counterfactual policies}: agents can recommend unseen items and receive immediate feedback. Simulators like Virtual-Taobao and KuaiSim show that exploration-heavy policies can improve real-world outcomes when paired with safeguards like action-norm constraints~\citep{ShiVirtualTaobao2018,ZhaoKuaiSim2023}. RecSim and RecSim NG allow controlled manipulation of user behavior, helping isolate genuine discoveries from exploiting simulator artifacts~\citep{IeRecSim2019,MladenovRecSimNG2021}. RecoGym bridges organic browsing and bandit interactions to analyze how exploration impacts click-through rates~\citep{RohdeRecoGym2018}, while the concept-drift generator introduces evolving data streams to test adaptive strategies~\citep{CaropreseModelling2025}. Collectively, these environments \textit{alleviate exploitation bias} by rewarding novel actions and modeling dynamic feedback.

Simulation thus offers a practical and rigorous way to study exploration--exploitation trade-offs in CMABs. The aforementioned frameworks span a spectrum from abstract (RecSim) to data-driven (KuaiSim) and probabilistic (RecSim NG), and their suitability depends on the research goal, such as latent-state discovery, session-level retention, or adaptation to concept drift. Future directions include adding timestamp semantics, multi-objective rewards, fairness modeling, and combining concept-drift generators with interactive simulators to assess bandit performance under realistic, non-stationary conditions.

\section{Key Takeaways and Future Directions}

Based on our analysis, we summarize below the key limitations of offline evaluation methodologies in the context of CMABs, as well as promising directions for addressing these shortcomings.

\begin{itemize}
\item Problems with current offline evaluation:

\begin{itemize}
    \item \textit{Bias toward exploitation:} offline evaluation systematically favors greedy or low-exploration strategies;
    \item \textit{Hyperparameter tuning distortions:} fine-tuning often converge toward configurations that reduce exploration;
    \item \textit{Off-policy evaluation remains limited:} even robust estimators failed to avoid exploitation bias;
    \item \textit{Limitations of simulators:} current simulators are often domain-specific and fail to model complex user behavior.
\end{itemize}

\item  Research directions to address these gaps:

\begin{itemize}
    \item \textit{Next-generation simulators:} design interactive environments with richer dynamics (e.g., non-stationarity, diverse item catalogs) to test exploration strategies meaningfully;
    \item \textit{Advancing OPE methods:} standardize off-policy evaluation techniques specifically tailored for complex recommender system environments, and build more datasets to support exploration-focused OPE estimators;
    \item \textit{Standardized benchmarks:} establish protocols that assess exploration beyond short-term accuracy;
    \item \textit{Broader metric reporting:} improve diversity metrics to provide a more holistic view of algorithm performance.
\end{itemize}
\end{itemize}

\section{Conclusions and Future Work}

This paper presented an extensive empirical comparison of linear Contextual Multi-Armed Bandits (CMABs) against a purely greedy linear regression baseline within offline evaluation settings. Our findings provide large-scale evidence that reinforces and quantifies the well-documented limitations of offline evaluation, particularly for assessing exploration strategies. Across a diverse range of datasets, the purely greedy linear model achieved superior or comparable accuracy to exploratory CMABs in 89\% of scenarios, underscoring this inherent bias towards exploitation. Furthermore, we observed that even when exploratory models aimed to enhance diversity, their performance in terms of accuracy often suffered, and hyperparameter optimization frequently converged towards settings that minimized exploratory behavior.

Despite these drawbacks, offline evaluation using historical logs persists as the predominant paradigm for assessing MABs, largely due to its operational simplicity and low cost. The clear disconnect between the long-term benefits of exploration and its often diminished performance in standard offline settings underscores a pressing need for the development and widespread adoption of more robust evaluation methodologies for MABs. Progress in simulation design, off-policy estimators, and standardized benchmarks will be key to advancing the state of interactive learning in recommender systems, ensuring fair algorithmic comparisons and ultimately delivering better user experiences.

\begin{acks}
This study was financed, in part, by the Brazilian Agencies CNPq (grant \#311867/2023-5), CAPES (Finance Code 88887.854357/2023-00), and FAPESP (Process Numbers \#2021/14591-7, \#2023/00158-5, and \#2024/15919-4).
\end{acks}

\bibliographystyle{ACM-Reference-Format}
\bibliography{references}


\end{document}